\documentclass{article}
\usepackage{amsmath,graphicx,mlspconf}
\usepackage{extarrows}
\usepackage{amsfonts,amssymb}
\usepackage{siunitx}
\newcommand{\vect}[1]{\mathbf{#1}}

\newcommand{\matt}[1]{\mathsf{#1}}
\toappear{2024 IEEE International Workshop on Machine Learning for Signal Processing, Sept.\ 22--25, 2024, London, UK}

\title{Ev-GS: Event-based Gaussian splatting for Efficient and Accurate Radiance Field Rendering}
\name{Jingqian Wu, Shuo Zhu, Chutian Wang, Edmund Y. Lam* \thanks{* Corresponding author: elam@eee.hku.hk}}
\address{The University of Hong Kong, Pokfulam, Hong Kong SAR, China}

\begin{document}

\maketitle

\begin{abstract}
Computational neuromorphic imaging (CNI) with event cameras offers advantages such as minimal motion blur and enhanced dynamic range, compared to conventional frame-based methods. Existing event-based radiance field rendering methods are built on neural radiance field, which is computationally heavy and slow in reconstruction speed. Motivated by the two aspects, we introduce Ev-GS, the first CNI-informed scheme to infer 3D Gaussian splatting from a monocular event camera, enabling efficient novel view synthesis. Leveraging 3D Gaussians with pure event-based supervision, Ev-GS overcomes challenges such as the detection of fast-moving objects and insufficient lighting. Experimental results show that Ev-GS outperforms the method that takes frame-based signals as input by rendering realistic views with reduced blurring and improved visual quality. Moreover, it demonstrates competitive reconstruction quality and reduced computing occupancy compared to existing methods, which paves the way to a highly efficient CNI approach for signal processing. 
\end{abstract}
\begin{keywords}
Event Camera, Radiance Field Rendering, 3D Gaussian splatting, Computational Neuromorphic Imaging
\end{keywords}
\section{Introduction}
\label{sec:intro}
The radiance field rendering aims to output the representation of the distribution and intensity of light across 3D space~\cite{klenk2023nerf}. With advancements in deep learning frameworks and CUDA-based rasterization, this rendering task has become vital for generating dense, photorealistic renderings of scenes in a 3D-consistent manner, playing a pivotal role in both computer vision and computer graphics  \cite{kerbl20233d}. Efforts have been directed towards curating real-scene datasets, often reconstructed from photographs or scans, to serve as benchmarks for evaluating algorithm performance on real-world data \cite{yang2023deformable}. However, real-world images captured by frame-based cameras frequently suffer from motion blur, caused by rapid motion and prolonged exposure duration \cite{klenk2023nerf, hidalgo2022event}. This phenomenon occurs when each pixel of the camera integrates light from different points in the scene during exposure, resulting in color value blending. Motion blur leads to a loss of information, hindering tasks such as radiance field rendering and subsequent processing \cite{szeliski2022computer}. Simply reducing exposure time to mitigate motion blur is often impractical, as it compromises light reception and exacerbates noise levels \cite{debevec2000acquiring}. Additionally, conventional frame-based cameras exhibit limited dynamic range, causing bright areas to saturate with white and dark areas to lose detail, particularly crucial information like text \cite{jung2010real}.

In contrast, event cameras capture asynchronous brightness changes per pixel, offering advanced properties such as high temporal resolution, high dynamic range, lower power consumption, and reduced latency compared to conventional frame-based cameras \cite{zuo2022devo}. 
Computational neuromorphic imaging (CNI) is a novel paradigm to harness the advanced properties of event cameras for numerous applications\cite{zhu2024computational}. 
In this context, to utilize the advantage of the event camera, we proposed the first CNI-informed method for inferring 3D Gaussian splatting (GS) from a monocular event camera, enabling efficient and accurate novel view synthesis for objects in grayscale space. Specifically, Our approach, named Ev-GS, leverages the advantages by employing 3D Gaussians as a flexible and efficient representation with purely event-based signal supervision, allowing accurate representation of challenging scenes for traditional frame-based supervision (e.g., motion blur, lacked frame number under high-speed movements, or insufficient lighting), while achieving high-quality rendering through faster training and real-time performance, especially for complex scenes and high-resolution outputs. Experiments show that our Ev-GS approach surpasses frame-based cameras by rendering realistic views with reduced blurring and improved visual quality on real-world datasets. Moreover, compared to existing rendering methods on synthetic datasets, Ev-GS demonstrates competitive rendering quality and significant efficiency improvements, including real-time rendering speed, memory occupancy, and training cost. We summarize our technical contribution as follows:
\vspace{-0.1cm}
\begin{itemize}
\item We introduce EV-GS, marking the first CNI-informed attempt to infer 3D GS from a monocular event camera, which enables a much more efficient synthesis of novel views for objects within the grayscale space compared to other state-of-the-art rendering methods. 
\vspace{-0.1cm}
\item We propose a novel event stream utilization and supervision framework specifically designed for differentiable 3D Gaussian-like methods for accurate and realistic scene rendering.
\end{itemize}

\section{Related Work}
\subsection{Neural Rendering and Radiance Field}
Recently, 3D GS \cite{kerbl20233d} has emerged as a compelling alternative to Neural Radiance Field (NeRF) \cite{mildenhall2021nerf} for 3D representation, exhibiting notable quality and speed improvements across both 3D and 4D reconstruction tasks. Its efficient differentiable rendering implementation and model design streamline training processes without necessitating spatial pruning \cite{kerbl20233d}. Despite its various advantages, a significant challenge lies in acquiring comprehensive and accurate scene information efficiently for training a 3D Gaussian model. Many approaches \cite{kerbl20233d, yan2024street, li2023animatable} adopt scene collection methods by capturing videos from a moving camera. This method offers the advantage of efficiently capturing training data: instead of capturing photos from every angle or direction, which can be time-consuming and prone to overlooking information, a single camera mounted on a moving device traverses the entire scene, ensuring all necessary information is captured. However, using frame-based cameras for this purpose may introduce issues such as motion blur and sparse frames, leading to deficiencies in view rendering results \cite{klenk2023nerf, hidalgo2022event}. To address this challenge, we present a new method to learn 3D Gaussian scene representations from event streams. It enables dense photorealistic novel view synthesis, overcoming the limitations associated with frame-based cameras.

\subsection{Neuromorphic Radiance Field Rendering}
The distinctive features of event cameras, including their capacity to prevent motion blur, provide a high dynamic range, ensure low latency, and consume minimal power, have spurred their adoption in the realm of computer vision and computational imaging with vital applications \cite{zhang2024joint, zhu2024efficient, wang2024neuromorphic, zhu2024computational}. Several approaches have been proposed to address the view synthesis challenge using NeRF \cite{mildenhall2021nerf} with event data \cite{klenk2023nerf, rudnev2023eventnerf}, leveraging volumetric rendering with either pure event or semi-event (blurred RGB involved) supervision. However, a significant drawback is the time-consuming optimization of NeRF. 
The computational demands of training and optimizing an event NeRF pipeline, in terms of both training time and GPU memory, are roughly 83 times more than the 3D GS pipeline. Moreover, the utilization of high-dimensional multilayer perception networks in the NeRF architecture results in a slower view-rendering speed of around 190 times, which may pose limitations for real-time rendering applications.

In this study, to the best of our knowledge, we proposed an adaptation of 3D GS \cite{kerbl20233d} to address view synthesis challenges, marking the first instance of such an application in this context. Our approach achieved efficient training and rendering processes while upholding realistic visual quality.

\section{Method}
	
	
	
	
	

\begin{figure*}[t]
\begin{center}
  \includegraphics[width=1 \linewidth]{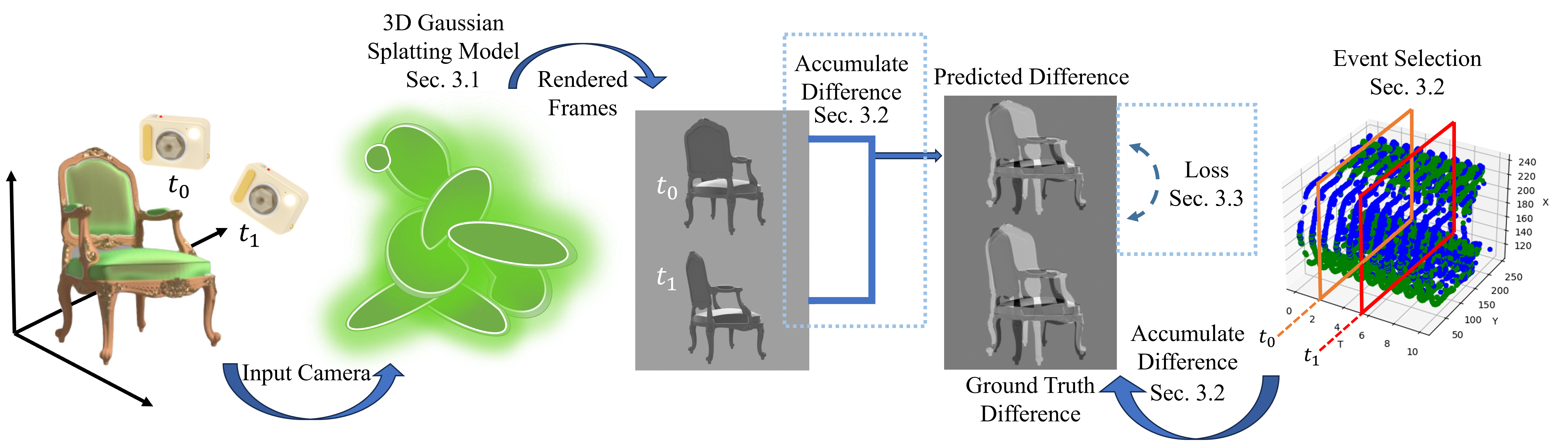}
\end{center}
   \caption{An overview of Ev-GS: a novel method for learning the radiance field volume from a moving event camera via 3D GS (Section \ref{3D GS}). We establish a link between the observed events and the rendered views at two distinct timestamps, $t_0$ and $t_1$, utilizing an event-based integral (Section \ref{Utilization}), under the supervision of pure event signal (Section \ref{Supervision}).}
\label{fig:pipeline}
\end{figure*}

\subsection{Preliminary on 3D Gaussian Splatting}
\label{3D GS}
3D GS \cite{kerbl20233d} portrays a detailed 3D scene by utilizing point clouds, with Gaussians utilized to delineate the scene's structure. In this depiction, each Gaussian is characterized by a central point, denoted as $\vect{x}$, and a covariance matrix $\Sigma$. The central point $\vect{x}$ is commonly referred to as the mean value of the Gaussian

\begin{equation}
    G(x) = \exp \left(-\frac{1}{2}\vect{x}^{T}\Sigma^{-1}\vect{x} \right).
    \label{GS_eq1}
\end{equation}

For the purpose of differentiable optimization, the covariance matrix $\Sigma$ can undergo decomposition into a rotation matrix $R$ and a scaling matrix $S$   
\begin{equation}
    \Sigma = RSS^TR^T.
\end{equation}

To generate renderings from different perspectives, the technique of splatting, as outlined in \cite{yifan2019differentiable}, is employed to position the Gaussians on the camera planes. This method, initially introduced in \cite{zwicker2001surface}, entails a viewing transformation denoted by $\matt{W}$ and the Jacobian $\matt{J}$ of the affine approximation of the projective transformation. Utilizing these parameters, the covariance matrix $\Sigma'$ in camera coordinates is

\begin{equation}
    \Sigma' = J W ~\Sigma ~W ^{T}J^{T}.    
\end{equation}

In summary, each Gaussian point within the model is defined by a collection of attributes: its position, represented by $\vect{x} \in \mathbb{R}^3$, its color depicted by spherical harmonics coefficients $\vect{c} \in  \mathbb{R}^k$ (where $k$ represents the degrees of freedom), its opacity $\alpha \in  \mathbb{R}$, a rotation quaternion $\vect{q} \in  \mathbb{R}^4$, and a scaling factor $\vect{s} \in  \mathbb{R}^3$. Specifically, for each pixel, the color and opacity of all Gaussians are computed based on the Gaussian representation outlined in Equation \ref{GS_eq1}. The blending process for color $C$ of N-ordered points overlapping a pixel adheres to a precise formula
\begin{equation}
    C = \sum_{i \in N} \vect{c}_i \alpha_i \prod_{j=1}^{i-1} (1-\alpha_j),
\end{equation}
where variables $\vect{c}_i$ and $\alpha_i$ denote the color and density of a specific point, respectively. These values are influenced by a Gaussian with a covariance matrix $\Sigma$, which is subsequently adjusted by adjustable per-point opacity and spherical harmonics color coefficients.

\subsection{Event Stream Utilization}
\label{Utilization}
Each event $e_k$ is described as a tuple $(\vect{x}_k, t_k, p_k)$, which occur asynchronously at pixel $\vect{x}_k = (x_k, y_k)$ at micro-second timestamp $t_k$. 

The polarity $p_k \in \{ -1 , + 1 \}$ denotes either an increase or decrease in the logarithmic brightness $L(\vect{x}_k, t_k)$ by the contrast threshold $A$. In other words, an event at time $t_k$ is triggered if the following condition is met

\begin{equation}
    \Delta L \triangleq L(\vect{x}_k, t_k) - L(\vect{x}_k, t_{k-1}) = p_kA,
\end{equation}
where $t_{k-1}$ is the timestamp that the previous event occurs at pixel $u_k$. 

As described in Figure \ref{fig:pipeline}, our goal is to render radiance field representation from differentiable 3D Gaussian functions under pure event signal supervision, with no RGB or grayscale frame-based data involved. To achieve this goal, we have to formulate the ground truth event data to a differentiable supervision signal and train a 3D GS model to render such representation. The core principle behind our algorithm is to generate two rendering results at two different camera poses on two timestamps, supervised by the ground truth event signal, which is the accumulated event frame between those two timestamps.

Specifically, we set a max window length of $W = 50$, and randomly select a window length of $w \in\mathrm{range}(W)$. For each timestamp $t$, we calculate two associated rendering result from the 3D Gaussian model $I_t = G(c_t)$ and $I_{t-w} = G(c_{t-w})$, where $I_t$ and $I_{t-w}$ are the rendered grayscale frame at time $t$ and $t-w$, $G$ is the 3D GS model, and $c_t$ and $c_{t-w}$ are the camera pose at time $t$ and $t-w$. We represented its logarithmic image as $L(I_t) = \log\big((I_t)^g + \epsilon\big)$, where $\epsilon = 1\times 10^{-5}$, 
 and $g$ denotes a fixed gamma correction value set to $4.8$ across all experiments. It conforms to the grayscale gamma curve, with Gamma $4.8$ serving as the recommended smooth approximation \cite{international1999multimedia}. As a result, we derive the predicted accumulative difference $E_{pred} = L(I_t) - L(I_{t-w})$.

On the other hand, to utilize event data and formulate it to a supervision signal for $E_{pred}$, following \cite{zihao2018unsupervised}, we aggregate the polarities of all events that transpire between the selected time $t$ and $t-w$ based on their positional information $u$, such that
\begin{equation}
E_{gt} = \int_{t}^{t+w} e_{k}(\vect{x}_{k}, p_{k}, t_{k}) \, dt,
\end{equation}
where $E_{gt}$ is the aggregated result.

\subsection{Event Stream Based Supervision}
\label{Supervision}
To effectively supervise $E_{pred}$ from $E_{gt}$, following v2e \cite{delbruck2020v2e} and \cite{klenk2023nerf}, we apply the linlog mapping described to derive the predicted logarithmic brightness difference for both $E_{pred}$ and $E_{gt}$. With a normalized $L_2$ loss, the loss $L_{e}$ can be calculated as 
\begin{equation}
    L_{e}(x, y) = \frac{\sum\limits_{i=1}^H \sum\limits_{j=1}^W (\| L(x) \|^2_2 - \| L(y) \|_2^2)^2}{H \times W},
\end{equation}
where for an arbitrary $u$

\begin{equation}
    L(u) = \mathop{\mathrm{linlog}} \big(I(u)\big) \triangleq
    \begin{cases}
     I(u) \times \ln(B) / B & I(u) < B \\
    \ln\big(I(u)\big)
    & \text{otherwise.}\\
    \end{cases}
\end{equation}


The threshold $B$ delineates the linear region, where no logarithmic mapping is applied. The value of $20$ is used for $B$ in all our experiments. $x$ and $y$ are $E_{pred}, E_{gt}$ correspondingly.

We also kept the D-SSIM loss used in the original 3D GS article, as we found a small coefficient would improve the rendering result. Following previous works, we calculate the D-SSIM loss as 
\begin{equation}
    L_{SSIM}(x, y) = \frac{(2\mu_{x}\mu_{y} + c_1) (2\sigma_{xy} +c_2)}{(\mu_{x}^2 + \mu_{y}^2 + c_1) (\sigma_{x}^2 + \sigma_{y}^2 + c_2)},
\end{equation}
where $\mu$ is the pixel sample mean, $\sigma_x$ is the variance of x, $\sigma_{xy}$ is the covariance of $x$ and $y$, $c_1 = (k_1L)^2$, $c_2 = (k_2L)^2$ are two variables to stabilize the division with weak denominator, L is the window range of the pixel values, and $k_1$, $k_2$ are constants. $x$ and $y$ are $E_{pred}, E_{gt}$ correspondingly. Therefore, the final loss function can be derived as follows
\begin{equation}
    L_{total} = L_{e} + \lambda (1- L_{SSIM}),
\end{equation}
where $\lambda$ is $0.1$ for all our experiments.

\section{Experiments}
\subsection{Overview}
\label{Exp Overview}
We analyze a total of four synthetic sequences and four real sequences. For the synthetic sequences, we utilize the 3D models sourced from \cite{mildenhall2021nerf}. Each scene undergoes rendering for a duration of one second, depicting a 360-degree rotation of the camera around the object at \SI{1000}{fps}  (frames per second), resulting in 1000 RGB images, then turned into 1000 grayscale images. From these images, we generate the event stream using the model \cite{gehrig2020video}, with the corresponding camera intrinsics and extrinsic directly applied in our approach.

\begin{figure}[t]
	
	\centering
	
	\includegraphics[width=\linewidth,scale=0.8]{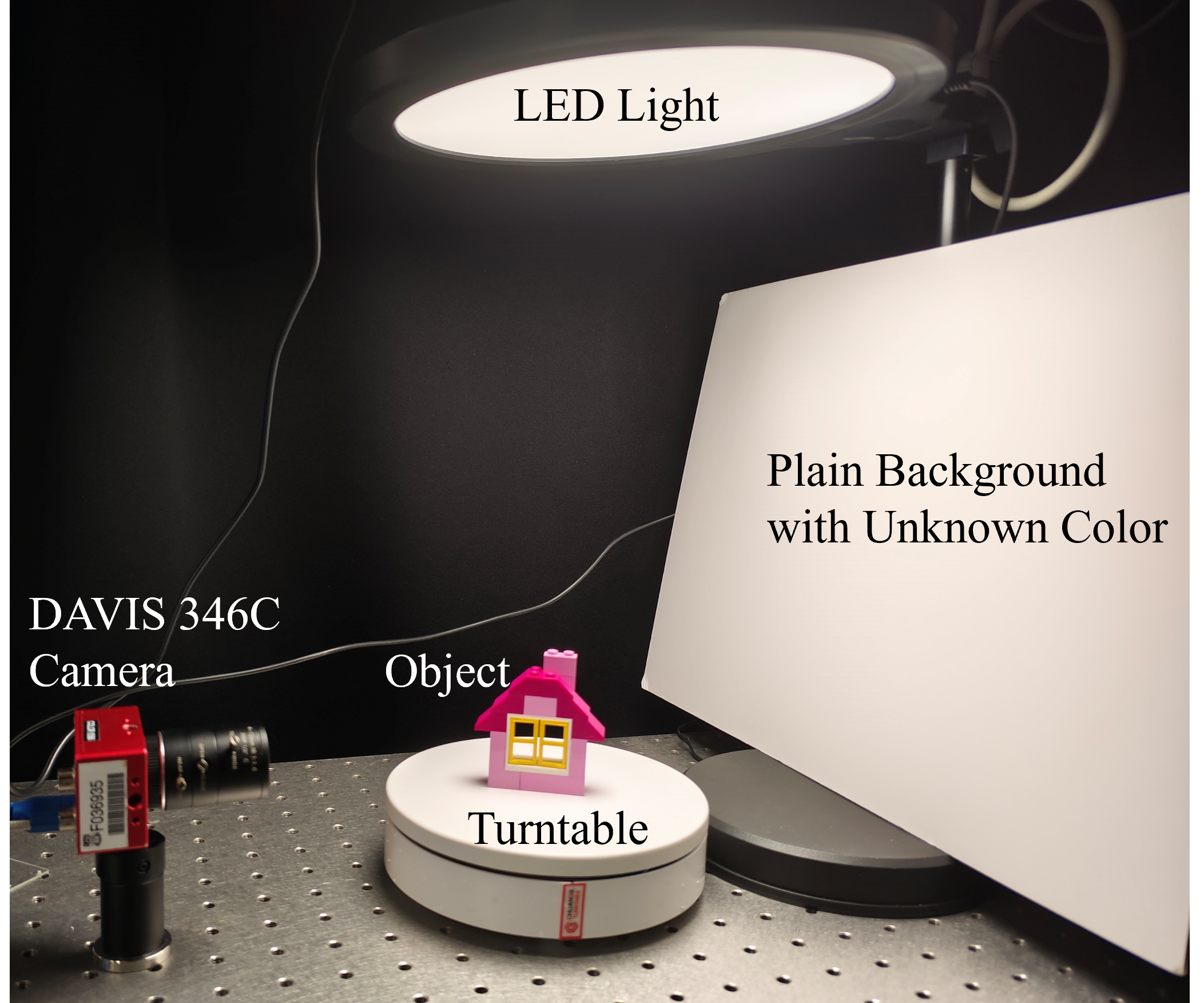}
	
	\caption{Demonstration of Hardware Setup.}
	
	\label{fig:setup}
	
\end{figure}

In our real-data experiments, we capture footage of four objects using the DAVIS 346C event camera. It is worth noting the challenges in creating a stable and calibrated setup in real-world scenarios where the event camera rotates around an object. To address this, we maintain the camera's static position and place the objects on a turntable with a maximum turning speed of five seconds per round. In this setup, maintaining constant lighting irrespective of the object's rotation angle is crucial. To achieve this, we mount a single USB ring light above the object. A photograph of the setup is provided by Figure~\ref{fig:setup}. 

The evaluation metrics for the rendering results of 3D GS include the peak signal-to-noise ratio (PSNR) and the structural similarity index (SSIM). PSNR measures signal fidelity by comparing the maximum signal power to noise power, while SSIM assesses image similarity based on luminance, contrast, and structure. These metrics provide quantitative insights into the quality and fidelity of the rendered images.

We further conduct an ablation experiment that demonstrates the necessity of our method design.

\subsection{Implementing Details}
Our implementation is constructed upon the 3D GS codebase \cite{kerbl20233d}, leveraging its foundational structure and functionalities. Because 3D GS requires a point cloud as input, we randomly initialized $10^6$ points to form the initial point cloud since the structure-from-motion \cite{schonberger2016structure} initialization, which is originally used in the model, is not applicable for event data. All experiments detailed in this paper are conducted by utilizing the computational resources of an NVIDIA RTX 3090 GPU.

We follow the settings of several hyperparameters for performance and optimization. The total number of iterations during the training process is set to $50,000$. For position optimization, the initial and final learning rates are $1.6 \times 10^{-4}$ and $1.6 \times 10^{-6}$. Learning rates for feature, opacity, scaling, and rotation optimization are $2.5 \times 10^{-3}$, $5 \times 10^{-2}$, $5 \times 10^{-3}$, and $1 \times 10^{-3}$, respectively.

\subsection{Synthetic Sequences}
\begin{figure}[t]
	
	\centering
	
	\includegraphics[width=\linewidth,scale=1.00]{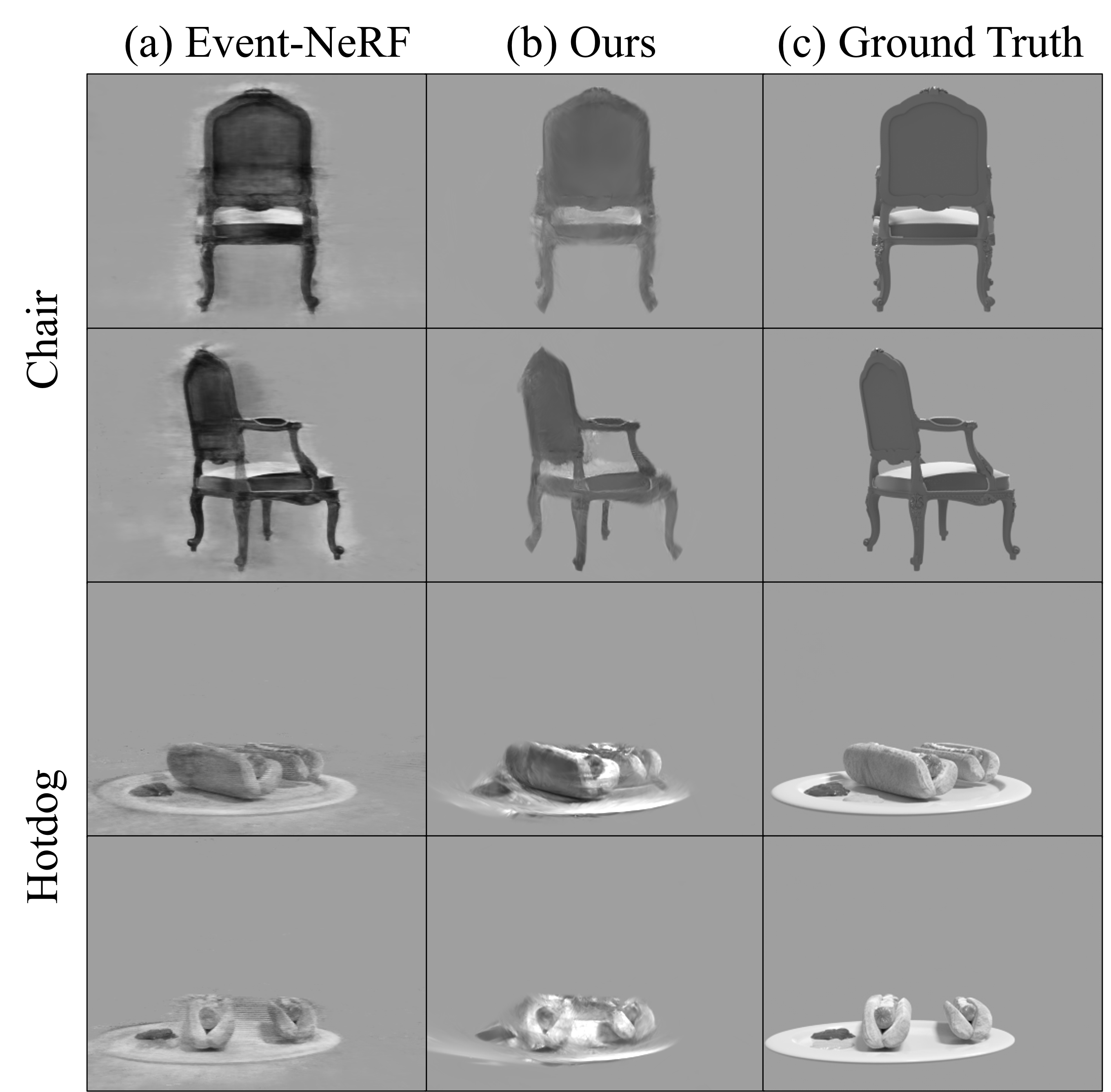}
	
	\caption{Visual Comparison between our approach and Event-NeRF.}
	
	\label{fig:vis synthesis}
	
\end{figure}
As previously mentioned, we evaluate synthetic sequences sourced from Mildenhall et al. \cite{mildenhall2021nerf}, encompassing various effects including chairs, hot dogs, ficus, and microphones. Table \ref{compare event nerf} demonstrates the quantitative comparison EventNeRF \cite{rudnev2023eventnerf}. According to our experiment results in Table \ref{compare event nerf}, our method demonstrates significant advantages over EventNeRF across various scenes. In most scenes, our approach achieves a higher PSNR and SSIM than EventNeRF, indicating superior reconstruction fidelity and better structural similarity between the reconstructed and ground truth images. Notably, our method boasts significantly reduced training times, taking only around 10 minutes compared to EventNeRF's $14$ hours for all scenes. This accelerated training time allows for more efficient model development and experimentation. Moreover, our method achieves higher temporal resolution, reaching around \SI{60}{fps} compared to EventNeRF's \SI{0.32}{fps}. A high frame rate is crucial for smooth and responsive visual experiences as it ensures real-time responsiveness, reduces motion sickness, and enhances productivity and competitiveness. Lastly, our method consumes less memory during the rendering stage. Ev-Gs utilizes only \SI{5}{GB} compared to EventNeRF's \SI{15}{GB}, which is advantageous for memory-constrained environments and facilitates scalability. 

Qualitatively, we also compare visual results for two sequences. As shown in Fig \ref{fig:vis synthesis}, our Ev-GS effectively learns view-dependent effects, structures, and intensity of the frame compared to the previous method.

\begin{table}[t]
\caption{Quantitative Comparison Against Event-NeRF \cite{rudnev2023eventnerf}.}

\scalebox{0.8}{
\begin{tabular}{c|ccccc}
\hline\hline
Metirc & PSNR↑ & SSIM↑ & Training Time↓ & FPS↑ & Memory↓ \\ \hline\hline
Scene & \multicolumn{5}{c}{Chair} \\ \hline
EventNeRF & 25.6 & 0.91 & 14h & 0.32 & 15GB \\
Ours & \textbf{28.1} & \textbf{0.93} & \textbf{9min} & \textbf{53.1} & \textbf{5GB} \\ \hline\hline
Scene & \multicolumn{5}{c}{Ficus} \\ \hline
EventNeRF & 27.1 & 0.91 & 14h & 0.33 & 15GB \\
Ours & \textbf{28.1} & \textbf{0.92} & \textbf{9min} & \textbf{63.0} & \textbf{5GB} \\ \hline\hline
Scene & \multicolumn{5}{c}{Hotdog} \\ \hline
EventNeRF & 26.0 & 0.92 & 14h & 0.32 & 15GB \\
Ours & 25.7 & \textbf{0.93} & \textbf{12min} & \textbf{55.3} & \textbf{5GB} \\ \hline\hline
Scene & \multicolumn{5}{c}{Mic} \\ \hline
EventNeRF & 25.0 & 0.915 & 14h & 0.32 & 15GB \\
Ours & 24.5 & \textbf{0.92} & \textbf{11min} & \textbf{66.02} & \textbf{5GB} \\ \hline\hline
\end{tabular}}
\label{compare event nerf}
\end{table}

\subsection{Real Sequences}
\begin{figure}[t]
	
	\centering
	
	\includegraphics[width=\linewidth,scale=1.00]{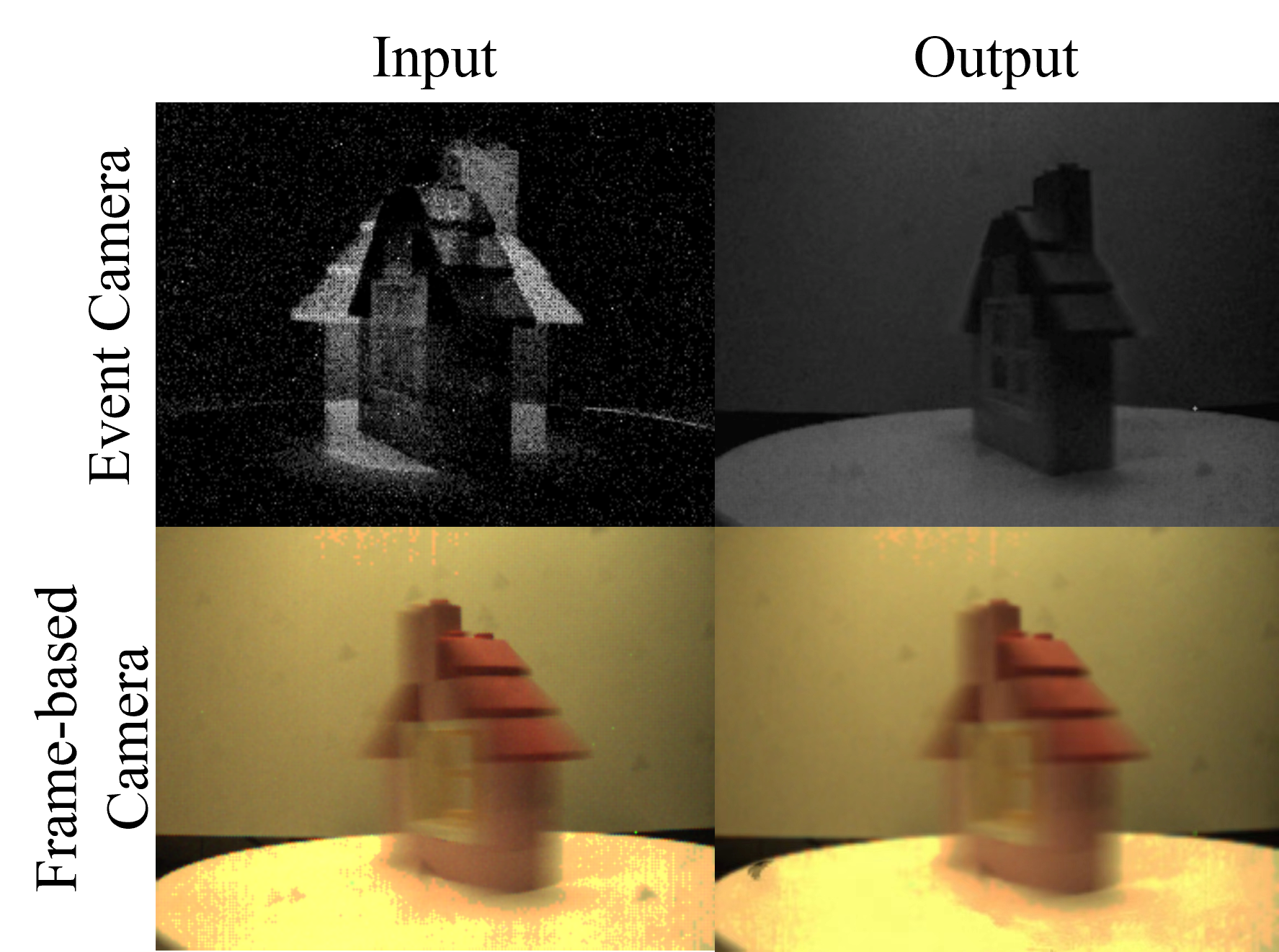}
	
	\caption{Visual Comparison of real-world data between event signal supervision and frame-based supervision.}
	
	\label{fig:real}
	
\end{figure}

As described in Section \ref{Exp Overview}, for real-world sequence, we analyzed a Lego toy scene captured on a turntable. Since no ground-truth RGB data is available, we assess the results visually, as depicted in Figure \ref{fig:real}. In scenarios with fast-moving cameras, traditional frame-based systems suffer from motion blur, resulting in blurry renderings. Our method, solely reliant on event data, circumvents motion blur and low-contrast issues, ensuring more precise and higher-quality rendering outcomes.

\vspace{-0.4cm}

\section{Conclusion}
In this work, we introduced Ev-GS, a novel method for inferring 3D Gaussian splatting from monocular event cameras. Leveraging the unique advantages of the CNI paradigm, Ev-GS enables efficient and accurate novel view synthesis in grayscale space. Our approach overcomes challenges by employing purely event-based supervision, resulting in superior rendering quality compared to existing methods. Experimental results demonstrate the effectiveness of Ev-GS in rendering realistic views with reduced blurring and improved visual quality on real-world datasets. Moreover, our method shows significant efficiency improvements, including real-time reconstruction speed and reduced memory occupancy, highlighting its potential for various applications within the CNI framework. However, Ev-GS still suffers from reconstructing difficult scenes, especially in complex objects with challenging textures. Future work will be done to fill the gap to enhance the robustness of reconstruction results both quantitatively and qualitatively. Overall, we believe Ev-GS will provide an enlightening reference and shed light on signal processing with event representation and its optical applications.

\vspace{-0.4cm}

\bibliographystyle{IEEEbib}
\bibliography{strings,refs}

\end{document}